\documentclass[wcp]{jmlr}

\usepackage{longtable}

\usepackage{booktabs}

\makeatletter
\def\set@curr@file#1{\def\@curr@file{#1}} 
\makeatother
\usepackage[load-configurations=version-1]{siunitx} 

\usepackage{times}
\usepackage{soul}
\usepackage{url}
\usepackage{hyperref}
\usepackage{inputenc}
\usepackage{graphicx}
\usepackage{amsmath}
\usepackage{booktabs}
\usepackage{algorithm}
\usepackage{algorithmic}
\jmlryear{}
\jmlrworkshop{CSCR I, 2020}

\title[]{A Vertical Federated Learning Method for Interpretable Scorecard and Its Application in Credit Scoring}

  \author{\Name{Fanglan Zheng} \Email{zhengfanglan@ebchinatech.com}\\
\Name{Erihe} \Email{erihe@ebchinatech.com}\\
\Name{Kun Li} \Email{likun3@ebchinatech.com}\\
  \Name{Jiang Tian} \Email{tianjiang@ebchinatech.com}\\
\Name{Xiaojia Xiang} \Email{xiangxiaojia@ebchinatech.com}\\
  \addr Everbright Technology CO. LTD, China
 }

\begin{document}

\maketitle

\begin{abstract}
With the success of big data and artificial intelligence in many fields, the applications of big data driven models are expected in financial risk management especially credit scoring and rating.
Under the premise of data privacy protection, we propose a projected gradient-based method in the vertical federated learning framework for the traditional scorecard, which is based on logistic regression with bounded constraints, namely FL-LRBC. The latter enables multiple agencies to jointly train an optimized scorecard model in a single training session. It leads to the formation of the model with positive coefficients, while the time-consuming parameter-tuning process can be avoided. Moreover, the performance in terms of both AUC and the Kolmogorov-Smirnov (KS) statistics is significantly improved due to data enrichment using FL-LRBC. 
At present, FL-LRBC has already been applied to credit business in a China nation-wide financial holdings group.
\end{abstract}


\section{Introduction}
The scorecard is a widely used assessment method in financial industry (refer to \cite{thomas2009consumer}), 
especially in risk management (refer to \cite{Raymond2007Credit}). The first purpose of the scorecard is to assist banks in making their credit lending decisions. It is no longer only applied in is applied in
assessing lending decisions, but also on-going credit risk
management and collection strategies.
Therefore, credit scoring is a tool to build a single aggregated risk indicator of a set of risk factors with a certain applicant. This tool should consist of a group of variables, 
statistically significant to be predictive in distinguishing between goods and bads.  

Typically, the scorecard model based on logistic regression (LR) is the most commonly used statistical technique for binary classification problem. 
It is widely used in banking industry and consuming finance due to its desirable features (e.g., interpretability and robustness). 
In order to further improve the prediction accuracy of the scorecard model,
two major schemes have been proposed. One is to train complex models such as 
ensemble trees or neural networks, since they are capable of learning 
non-linear decision boundary. Due to the lack of interpretability, however, 
it is difficult to apply to real credit assessment. The other is to enrich 
the data set in terms of sample size or characteristics. 
Also, it is challenged under the premise of Data Privacy Protection Act, 
since the transferring of private data is always involved.

Recently, federated learning (FL) has been proposed as a potential solution for 
data isolation at the cost of mass of training time, due to the relatively heavy
encryption and a large number of communication rounds. 
FL is an emerging frontier field studying privacy-preserving
collaborative machine learning while leaving data instances at their providers locally. 
It enables different agencies to collaboratively train a shared machine learning model 
without transferring any data from one data provider to another. 
Refer to \cite{yang2019federated}. In practice, to avoid variables' multi-collinearity 
in training a scorecard model, it inevitably spends plenty of time in processing with parameter-tuning. 
It makes the model coefficients greater than or equal to zero, so as to retain
the interpretability and robustness of the scorecard. Accordingly, it is a big
challenge to train an optimized model in the FL framework,
due to the intolerable machine time cost.

In order to address this problem, we propose a projected gradient-based method 
for LR with bounded constraints (FL-LRBC) in the FL framework (FATE, refer to  \cite{webank_fate_2018}).
Based on FL-LRBC, an optimized scorecard can be obtained in a single training session. 
It not only prevents from time-consuming parameter-tuning, but also keeps model interpretable. 
We further evaluate the performance of our credit
scorecard product on the large scale real-world data sets from
a nation-wide joint-stock commercial bank (BANK)
and 
a Cloud Payment company (CLOUD PAYMENT), which belong to
a China financial holdings group (GROUP).
Experimental results show that an average of test AUC of $72.6\%$ is achieved, 
which is about $9\%$ higher than that without FL-LRBC.

The rest of this manuscript is organized as follows. In Sect. \ref{lr},the related work about
LR with bounded constraint in the FL framework and its projected gradient-based method are discussed. 
Sect. \ref{er} gives the details of experimental results in our FL-LRBC framework. 
Conclusions are presented in Sect. \ref{con}.

\section{LR with Bounded Constraint in the Vertical FL Framework}\label{lr}
This section consists of two parts. First, we introduce the LR-based scorecard and weight of evidence, respectively. Second, we propose a projected algorithm for bound-constrained LR optimization in the vertical FL framework.

\subsection{LR-based Scorecard Model and Weight of Evidence}
The traditional scorecard model is the use of LR to translate relevant data into numerical scores that guide credit decisions. 
LR uses maximum likelihood estimation process, which transforms the 
dependent variables into a log function and estimates the regression coefficients
in a way that it maximizes the log-likelihood. 
Mathematically, it equals to solve a non-constrained optimization problem, 
which is formulated as follows:
\begin{equation}
\label{LR_UN}
\begin{split}
        \min_{\substack{\mathbf{w}\in\mathbf{R}^n\\
    b\in \mathbf{R}}}&\frac{1}{T}\sum_i^T\log(1+\exp(-y_i(\mathbf{w}^{\mathbf{T}} \mathbf{x}_i+b)),
    \end{split}
\end{equation}
yielding an occurrence probability and log odds, 
\begin{equation}\label{PROB}
    p_{i}=\frac{1}{1+e^{-(\mathbf{w}^{\mathbf{T}}\mathbf{x}+b)}}
\end{equation}
\begin{equation}\label{ODDS}
    \log(\mathrm{odds})=\log(\frac{p_{i}}{1-p_{i}})=(\mathbf{w}^{\mathbf{T}} \mathbf{x}+b)
\end{equation}
where $\mathbf{w}$, $\mathbf{x}_i$ and $y_i$ are coefficient vector, 
i-th feature vector and its corresponding label, respectively.
Here $p_{i}=p(y_i=1)$ and $b$ are the probability of $y_i=1$ and the intercept.

In the scorecard model, variables are seldom represented in their original form.
They are usually segmented into grouping intervals, with the aims of creating bins for the model that maximize the correlation with these variables. Variable transformation through binning is an essential part of LR model,
measured by weight of evidence (WOE). WOE impacts the predictive power of scorecard model,
and is often used as a benchmark to screen variables.
In the credit risk modeling, WOE converts the risk 
associated with a particular choice into a linear scale that is easier for the human mind to evaluate:
\begin{equation}\label{WOE}
    \rm WOE_i=\ln(\frac{Bad_i/\sum Bad_i}{Good_i/\sum Good_i})=
    \ln(\frac{Bad_i/ Good_i}{\sum Bad_i/\sum Good_i})
\end{equation}
here, $i$ is the $i$-th group/category or bin. The precondition is none-zero for all $\rm Bad_i$ and $\rm Good_i$. A negative WOE is that the proportion of defaults is higher for that attribute than the overall proportion, indicating higher risk. 
Besides, WOE does not suffer from the outliers and treats the missing values as a special group.

\begin{figure}[htbp]
\centering
\includegraphics[width=0.5\linewidth]{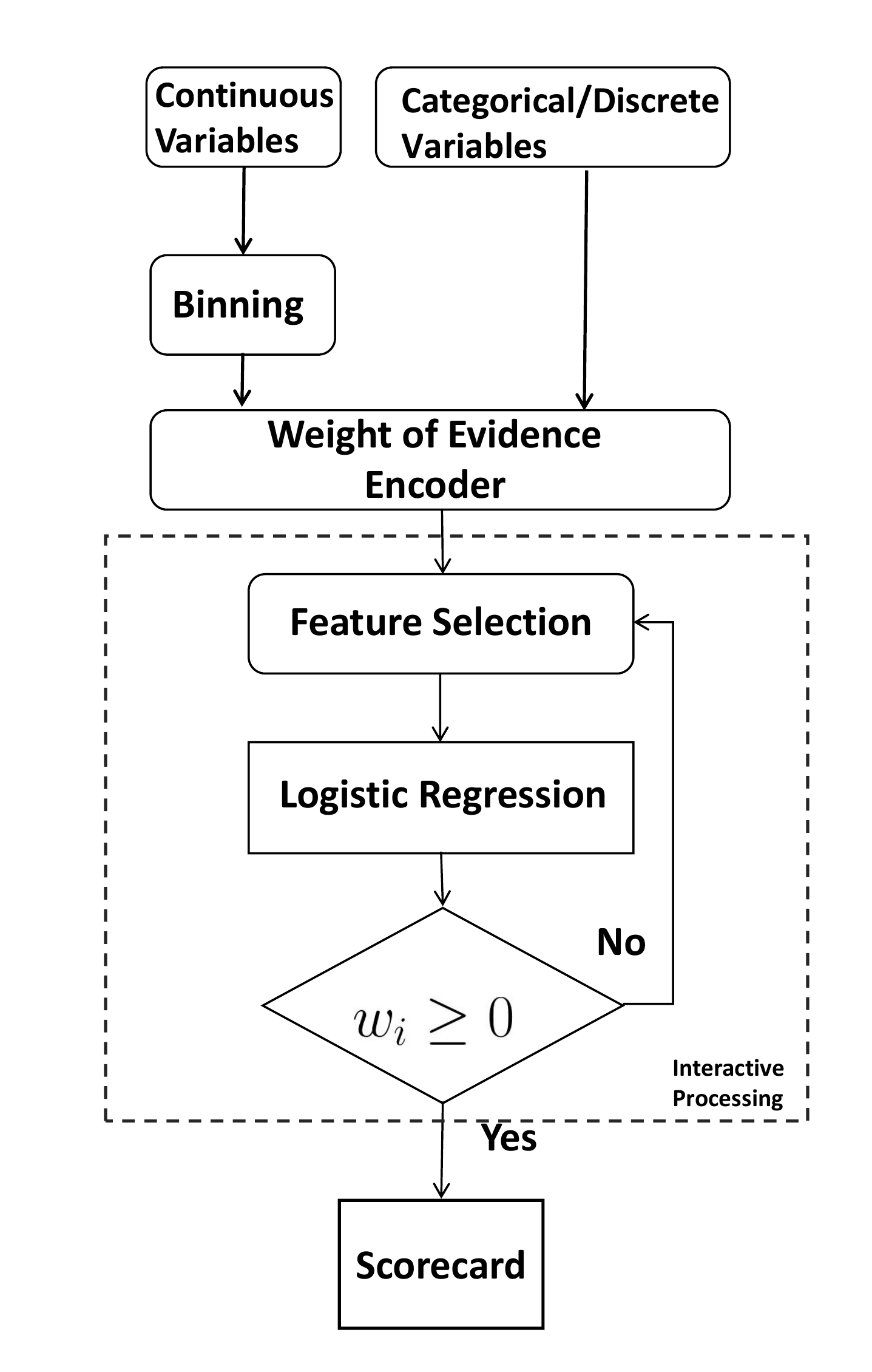}
\caption{The typical processing of a scorecard model building.}
\label{fig:sc_flow} 
\end{figure}.

For a WOE variable, it has a linear relationship with the logistic function, making it well suitable for representing its original form when using LR. 
In essence, WOE attempts to find a monotonic relationship between the input and target variables, which is "bad/good" account in scorecard and "events/non-events" or "1/0"("1/-1") in the general case. In short, WOE transformation helps to form a strict linear relationship with log odds.
In order to ensure interpretability and robustness of LR, the elements of weight vector $\mathbf{w}=(w_1,...,w_i,...,w_n)$ 
should be non-negative i.e.
\begin{equation}\label{wp}
w_i\ge 0.
\end{equation}
It is easily understood by taking variable $x_i$ in  bin (or category) $b_j$ with $\rm WOE_j>0$ as an example. 
That is, $\rm Bad_j/ Good_j>\sum Bad_j/\sum Good_j$ means higher bad rate than average level, which is consistent with higher odds. 
Otherwise, $w_ix_i$ for $b_j$ would imply the opposite effect with log odds, 
resulting in failure of model interpretability that lower log odds would coexist with higher bad rates.

Due to the multi-collinearity, however, model coefficients may not satisfy Eq.(\ref{wp}).
Usually, variable selection techniques with variance inflation factors (VIF) or correlation 
matrix are used to overcome this problem as well as achieve an optimized model with $w_i\ge 0$. This processing runs under the supervision of the model builder or follows some rules
in step-wise regression, which is referred as "Interactive Processing" in Figure \ref{fig:sc_flow}. 
According to our knowledge, the model builders in the risk management still follow this typical processing.

\subsection{A Projected Algorithm in the FL Framework for Bound-Constrained LR Optimization}
The vertical federated learning (VFL) framework for logistic regression was first introduced in \cite{aono2016scalable,hardy2017private}. It is based on \cite{Paillier1999} encryption and implemented in the open-source project for secure computing framework to support the federated AI ecosystem \cite{webank_fate_2018}. In this framework, even a single training session becomes
time expensive due to heavy encryption and communication for data privacy-protection.
It leads to the fact that the typical parameter-tuning manner to avoid multi-collinearity 
is no longer feasible. 
From a mathematical perspective, it evolves from a non-constrained into bound-constrained optimization problem,
\begin{equation}
\label{LR_OP_BD}
\begin{split}
        \min_{\substack{\mathbf{w}\in\mathbf{R}^n\\
    b\in \mathbf{R}}}&\frac{1}{T}\sum_i^T\log(1+\exp(-y_i(\mathbf{w}^{\mathbf{T}} \mathbf{x}_i+b))\\
    s.t. &\quad w_i\ge 0
    \end{split}
\end{equation}
whose solution is not available in the FL framework. 
Accordingly, we propose a projected gradient-based method as a solution.

For a general bound-constrained optimization problem, 
optimal conditions are related to sign condition on multipliers in KKT conditions \cite{Nocedal2006numerical}.
Several algorithms based on the gradient-based projection method, such as L-BFGS-B \cite{byrd1995a}, 
can be used to solve this bound-constrained optimization problem. 
In the following theorems, $l_i=0$ and $u_i=+\infty$ in the scorecard model.  

\begin{theorem}[Optimal Conditions for Bound Constraints]
Let $f(x)$ be continuously differentiable. If $x^*$ is the local minimizer of
\begin{align*}
        \min_{x\in\mathbf{R}^n} &\quad f(x)\\
    s.t. &\quad l\le x\le u
\end{align*}
then
\begin{align*}
\left( \frac{\partial f}{\partial x_i}\right)_{x=x^*}\left\{\begin{matrix}
 \ge 0,& if\, x_i^*=l_i \\ 
 =0,& if\, l_i<x_i^*<u_i \\ 
 \le 0,& if\, x_i^*=u_i  
\end{matrix}\right. .
\end{align*}
\end{theorem}
Use the "projection" operator defined as
\begin{gather*}
[P_{[l,u]}(x)]_i=\left\{\begin{matrix}
 l_i,& if\, x_i\le l_i \\ 
 x_i,& if\, l_i<x_i<u_i \\ 
 u_i,& if\, x_i\ge u_i  
\end{matrix}\right.
\end{gather*}
and the following first-order condition holds.
\begin{theorem}[First-Order Conditions for Bound Constraints]
Let $f(x)$ be continuously differentiable. If $x^*$ is local minimizer, then
\begin{align*}
x^*=P_{[l,u]}(x^*-\nabla f(x^*)).
\end{align*}
\end{theorem}

\begin{figure*}[htbp]
\centering
\includegraphics[width=0.85\linewidth]{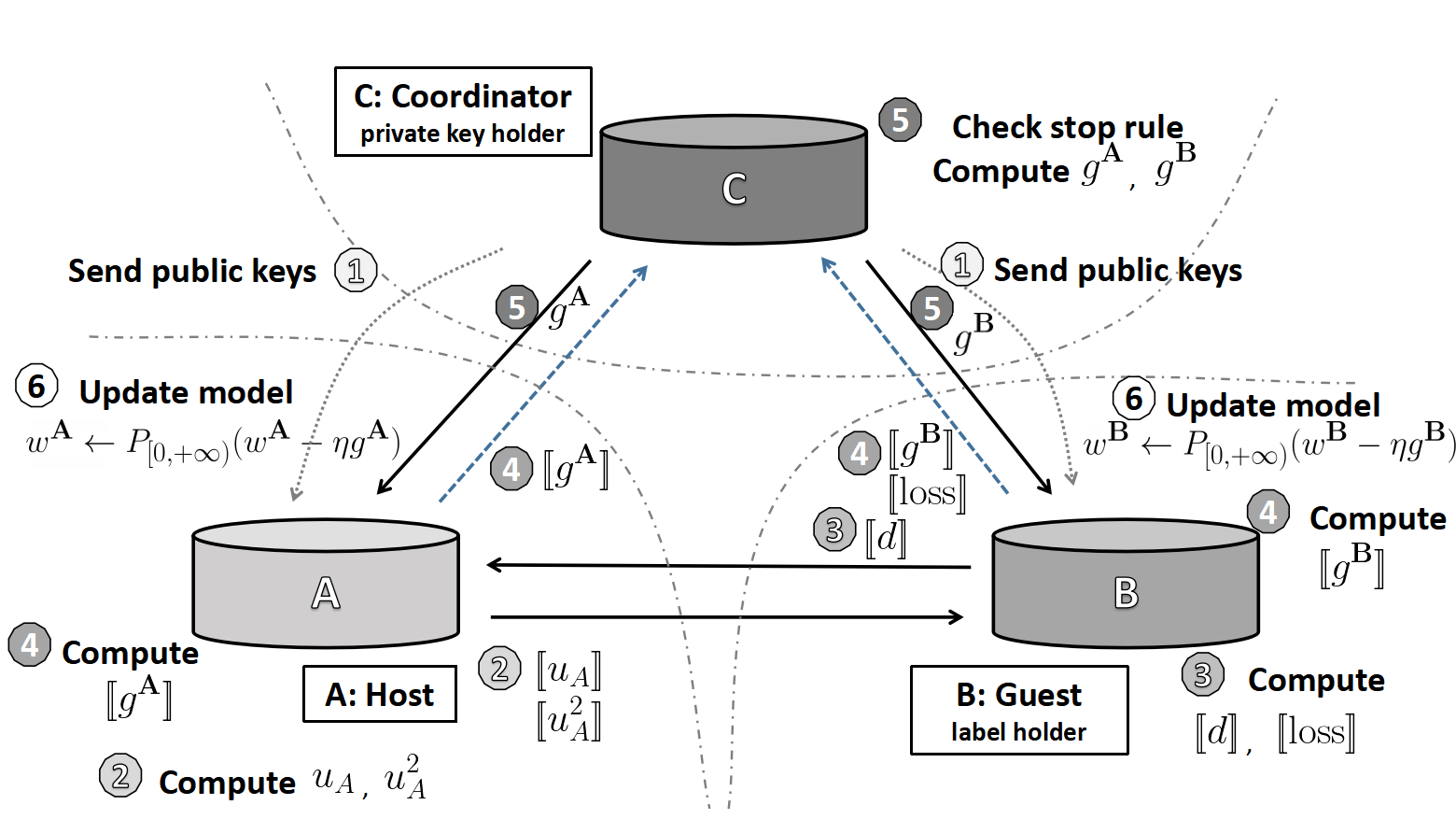} 
\caption{ A Framework of LR with Bounded Constrains in the vertical FL Framework.}
\label{fig:VFL} 
\end{figure*}

Different from the common optimization methods, such as stochastic gradient descent (SGD) and stochastic quasi-newton (SQN) methods \cite{yang2019Quasi}, 
we apply the "projection" operator as Eq.\ref{LR_OP_BD_PR} shows in each updating step of $w_i$ to achieve a model with $w_i\ge 0$.
\begin{align}\label{LR_OP_BD_PR}
\mathbf{w}_{k+1}=P_{[0,+\infty)}(\mathbf{w}_k-\eta \hat{g}_k)
\end{align}
where $\eta$ and $\hat{g}_k$ are learning rate and descent direction, respectively. 
Also, the initial value of $\mathbf{w}_i$ should belong to ${\mathbf{R}^n}^+$.
Here, the vertical federation learning is investigated. The original datasets are vertically partitioned and distributed on two honest-but-curious private \textbf{Party A} (the host data provider with only features $X^A\in\mathbf{R}^{T\times n_A}$) and \textbf{Party B} (the guest data provider with features $X^B\in\mathbf{R}^{T\times n_B}$ and labels $y$). Each party owns a disjoint subset of data features over a common sample IDs with $X = (X^A, X^B)$. The homomorphic encryption schemes such as \cite{Paillier1999} allow any party can encrypt their
data with a public key, while the private key for decryption is owned by the third party, i.e., the \textbf{coordinator}. With additively homomorphic encryption we can compute the additive of two encrypted numbers as well as the product of an unencrypted number and an encrypted one, which can be denoted as $[\![u]\!]+[\![ v]\!] = [\![ u + v]\!]
$, $v[\![ u]\!]= [\![ vu]\!]$ by using $[\![\cdot]\!]$ as the encryption operation.
The loss and gradient at \textbf{Party A\&B} are computed as follows. Let $\mathcal{S}\subseteq \{1,..., T\}$ be the index set of the chosen mini-batch data instances. The corresponding loss and gradient are given on $\mathcal{S}$. By denoting $u_A = \{u_A[i] =(w^A)^{\mathbf{T}}x_i^A:i\in\mathcal{S}\}$, $u_A^2=\{u^2_A[i] =((w^A)^{\mathbf{T}}x_i^A)^2:i\in\mathcal{S}\}$ for \textbf{Party A} (similarly $u_B$ and $u_B^2$ for \textbf{Party B}) and $d = \{d_i=\frac{1}{4}(u_A[i]+u_B[i])-\frac{1}{2}y_i: i\in \mathcal{S}\}$, the encrypted loss and gradient can be computed by transmitting $[\![u_A]\!]$
from  \textbf{Party A} to \text{Party B}, and transmitting $[\![d]\!]$ from \textbf{Party B} to \textbf{Party A} following 
\begin{equation}
\label{LR_formula_1}
\begin{split}
[\![\mbox{loss}]\!]\approx &\frac{1}{|\mathcal{S}|}\sum_{i\in\mathcal{S}}
\left[ [\![\log 2]\!]-\frac{1}{2}y_i([\![u_A[i]]\!]+[\![u_B[i]]\!])\right.\\
+&\left.\frac{1}{8}([\![u^2_A[i]]\!]+2u_B[i][\![u_A[i]]\!]+[\![u^2_B[i]]\!])\right]
\end{split}
\end{equation}

\begin{equation}
\label{LR_formula_2}
\begin{split}
[\![g]\!]\doteq& ([\![g^A]\!],[\![g^B]\!])\approx \frac{1}{|\mathcal{S}|}\sum_{i\in\mathcal{S}}
[\![d_i]\!]x_i \\
=&\left(\frac{1}{|\mathcal{S}|}\sum_{i\in\mathcal{S}}[\![d_i]\!]x^A_i,\frac{1}{|\mathcal{S}|}\sum_{i\in\mathcal{S}}[\![d_i]\!]x^B_i\right)
\end{split}
\end{equation}
here $$[\![d_i]\!]=\frac{1}{4}([\![u_A]\!]+[\![u_B]\!])+[\![-\frac{1}{2}y_i]\!]. $$

\begin{algorithm}[htbp]
\caption{A Framework of LR with Bound-Constraints in the Vertical FL Framework.}
\label{alg:algorithm}
\textbf{Input}: $w^A_0$,$w^B_0$\\
\textbf{Parameter}: $\eta$\\
\textbf{Output}: $w^A$,$w^B$
\begin{algorithmic}[1] 
\STATE Project $w^A_0$ and $w^B_0$ to ${\mathbf{R}^{n_A}}^+$ and ${\mathbf{R}^{n_B}}^+$,.
\WHILE{two steps difference condition (loss or coefficients) and max iteration number condition}
\STATE Choose a mini-batch $\mathcal{S}$
\STATE \textbf{Party A\&B:} Compute $[\![\mbox{loss}]\!]$, $[\![g]\!]$ as Eq.(\ref{LR_formula_1})(\ref{LR_formula_2})
\STATE \textbf{Coordinator:} Compute $g^A$, $g^B$ and $\mbox{loss}$ with private key and determine the stop condition. 
\STATE \textbf{Party A\&B:} Update each one's own model\\
$w^A\leftarrow P_{[0,+\infty)}(w^A-\eta g^A)$\\
$w^B\leftarrow P_{[0,+\infty)}(w^B-\eta g^B)$
\ENDWHILE
\STATE \textbf{return} $w^A$,$w^B$
\end{algorithmic} 
\end{algorithm}

More details about iterative flow of FL-LRBC can be seen in Figure.\ref{fig:VFL} and Algorithm.\ref{alg:algorithm}. In fact, the key point is the updating step (refer to Eq.(\ref{LR_OP_BD_PR})). It also can be applied to other gradient descent (GD) optimization algorithms and the quasi-newton method.

\section{Experimental Results}\label{er}
This section is organized with three parts. First, we introduce the data sets that are used in our FL-LRBC framework. 
Second, we show the performance measures in the scorecard model. Finally, we demonstrate the results of our experiments.

\subsection{Data Set Description}
For risk management in the retail credit business scenario, 
variable names of the relevant data in finance agencies won't be disclosed. 

With the usage of FL-LRBC, BANK trains a scorecard model with
CLOUD PAYMENT collaboratively, 
which has China's convenience payment data. 
There are 4,000,000 observations in the data set in all. 
Note that the target variable is coded as 1 and 0 to 
indicate default (according to a default definition chosen by ) and non-default, respectively. 

There are 464,454 defaults (events) and 3,535,546 non-defaults in the data set and no indeterminant is defined.
The total data set consists of 9 and 5 variables with information value (IV) more than a threshold 0.02 from the above two agencies, respectively. 
It can be seen from Table \ref{tab:plain1}.
The purpose of this is to filter out variables with strong predictive 
power that are input for training model.  

\begin{table}[htbp]
\centering
\begin{tabular}{llll}
\hline
variable  &data type &missing rate ($\%$)  &IV \\
\hline
var0      &float     & 49.3     & 0.320        \\
var1      &float     & 52.5     & 0.314      \\
var2      &char      & 0.0      & 0.312       \\
var3      &float     & 0.0      & 0.215       \\
var4      &char      & 0.0      & 0.211       \\
var5      &float     & 16.4     & 0.154        \\
var6      &float     & 0.0      & 0.108       \\
var7      &float     & 0.0      & 0.07       \\
var8      &float     & 0.0      & 0.055       \\
var9      &float     & 0.0      & 0.055       \\
var10     &char      & 0.0      & 0.052       \\
var11     &float     & 0.0      & 0.047       \\
var12     &char      & 0.0      & 0.046       \\
var13     &char      & 0.0      & 0.042      \\
\hline
\end{tabular}
\caption{Variables with IV more than 0.02 are sorted in the descending order.}
\label{tab:plain1}
\end{table}

Compared with the simple joint training, FL-LRBC maintains their models in both BANK and 
CLOUD PAYMENT without any data transferring, locally.

\subsection{Model Performance Measures}

Once a scorecard model completes, 
its practicality is often evaluated in terms of its predictive and discrimination ability. 
A complete description of predictive accuracy is given
by the Receiver Operating Characteristic (ROC) curve.
A model with high accuracy will have high sensitivity and specificity simultaneously, 
leading to an ROC curve which goes close to the top left corner of the plot. 
The area under the ROC curve, AUC, as one of the two main performance metrics, 
provides a measure of the model's predictive ability between those observations 
that experience the outcome of interest versus those who do not. 
This is called a summary measure of the accuracy of a quantitative diagnostic test. 
AUC is also referred to as the C statistic.

Kolmogorov-Smirnov (KS) statistic, the other performance metrics, 
is commonly used in measuring discrimination power of scorecard systems.
Also, KS curve is mostly used as a data-visualization tool to illustrate the model's effectiveness. 
The cumulative distributions of default and non-default are plotted against the score 
(output from the model, transformed, can also be seen as the different covariate patterns, ranked in a way). 
Geometrically, the maximum distance is then calculated between these two distribution curves,
which is defined as  
\begin{equation}\label{KS}
    \displaystyle
    \rm D_{KS}=\max\left|cp_1 - cp_0\right|
\end{equation}
where $\rm cp_1$ and $\rm cp_0$ are the cumulative percentage for defaults and non-defaults, separately.
Here $||$ indicates taking the absolute value and max is taken of all the differences over all possible scores. The larger $\rm D_{KS}$, the stronger the model’s discrimination power.

\subsection{Results and Discussions}
As for continuous variables, discretion can be accomplished via
 binning. 
The style of binning is always inspired by scorecard
construction methods. For categorical variables, 
no binning is needed and the histogram estimator can be used directly.
Then, the original variables are transformed into WOE variables (var\_{woe}).   

\begin{table}[htbp]
\centering
\begin{tabular}{ll}
\hline
variable  &coefficient  \\
\hline
intercept        &-2.85        \\
var0\_{woe}      &1.99            \\
var1\_{woe}      &0.58       \\
var3\_{woe}      &0.91      \\
var4\_{woe}      &0.95      \\
var6\_{woe}      &1.21     \\
var8\_{woe}      &0.83   \\
var9\_{woe}      &0.36   \\
var10\_{woe}     &0.19    \\
var13\_{woe}     &1.17    \\
\hline
\end{tabular}
\caption{Summary of the scorecard in our FL-LRBC framework.}
\label{tab:plain2}
\end{table}

To minimize the impact of over-fitting, we split the data set into $70\%$ training and $30\%$ testing parts. Using our FL-LRBC framework, the scorecard model is achieved in a single training session. Ina ll, We spend about 30 hours to obtain this optimized scorecard.
Otherwise, time cost would be more than hundreds of machine hours,
due to iterative feature-selecting and parameter-tuning.

\begin{figure}[htbp]
\centering
\includegraphics[width=0.45\textwidth]{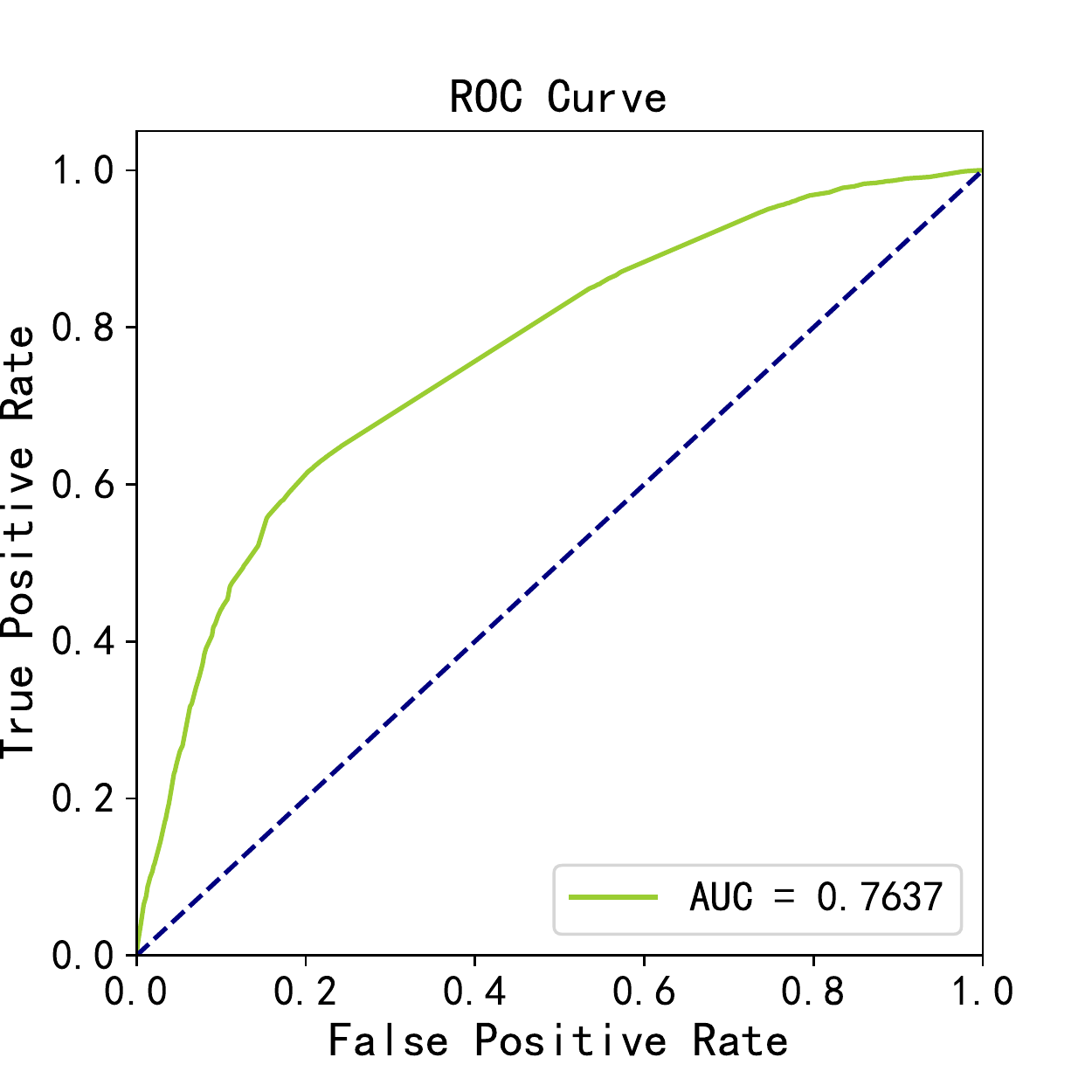}
\includegraphics[width=0.45\textwidth]{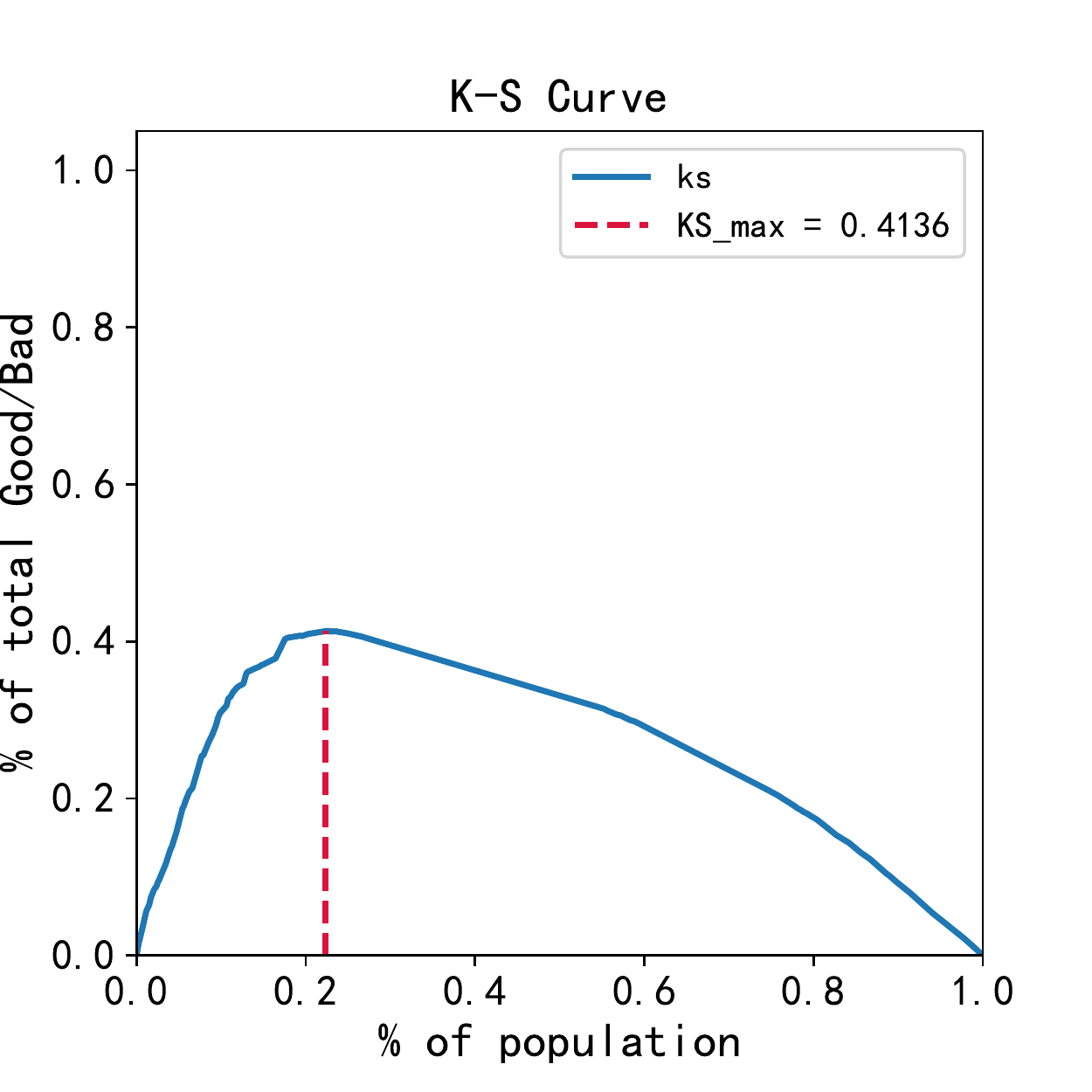}
\includegraphics[width=0.45\textwidth]{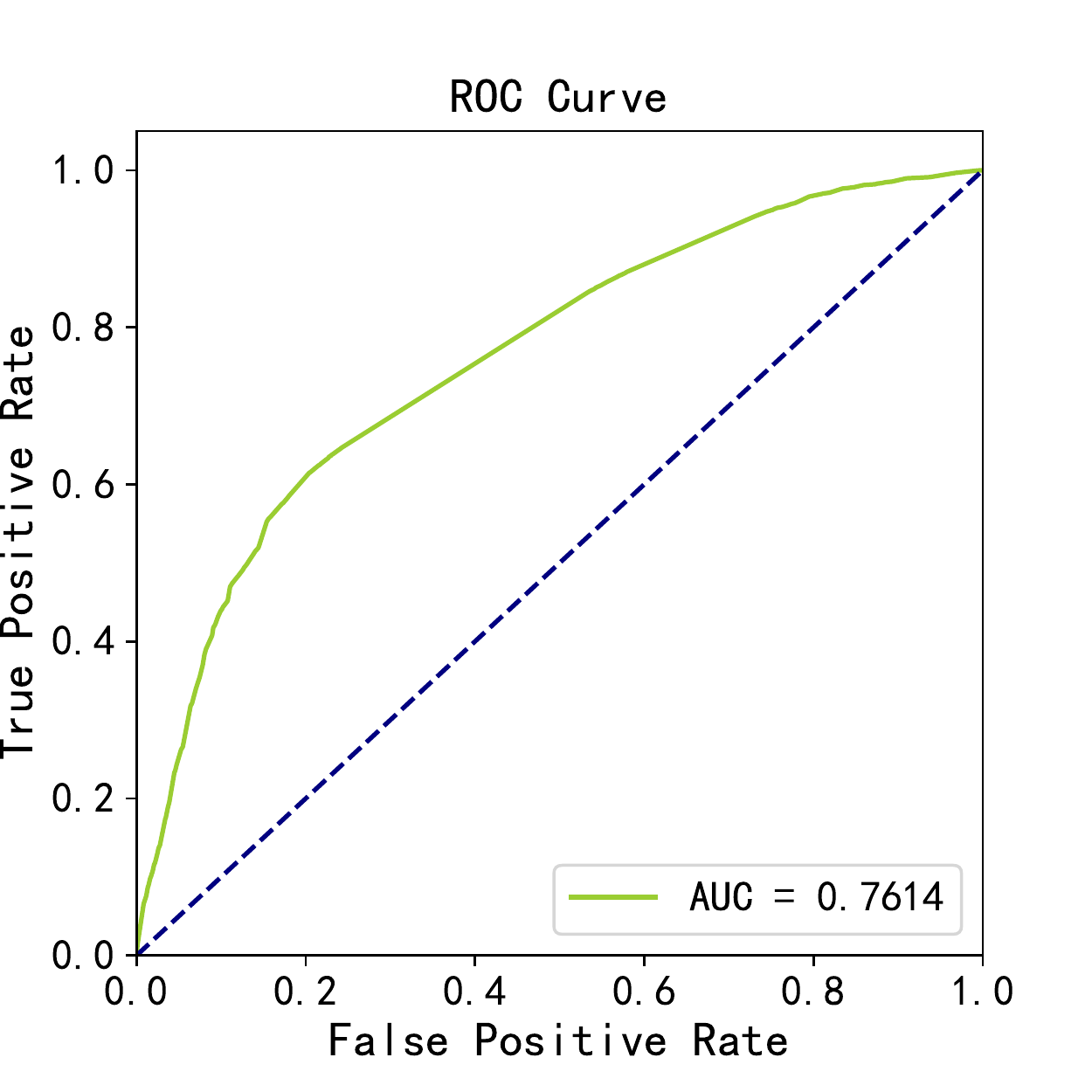}
\includegraphics[width=0.45\textwidth]{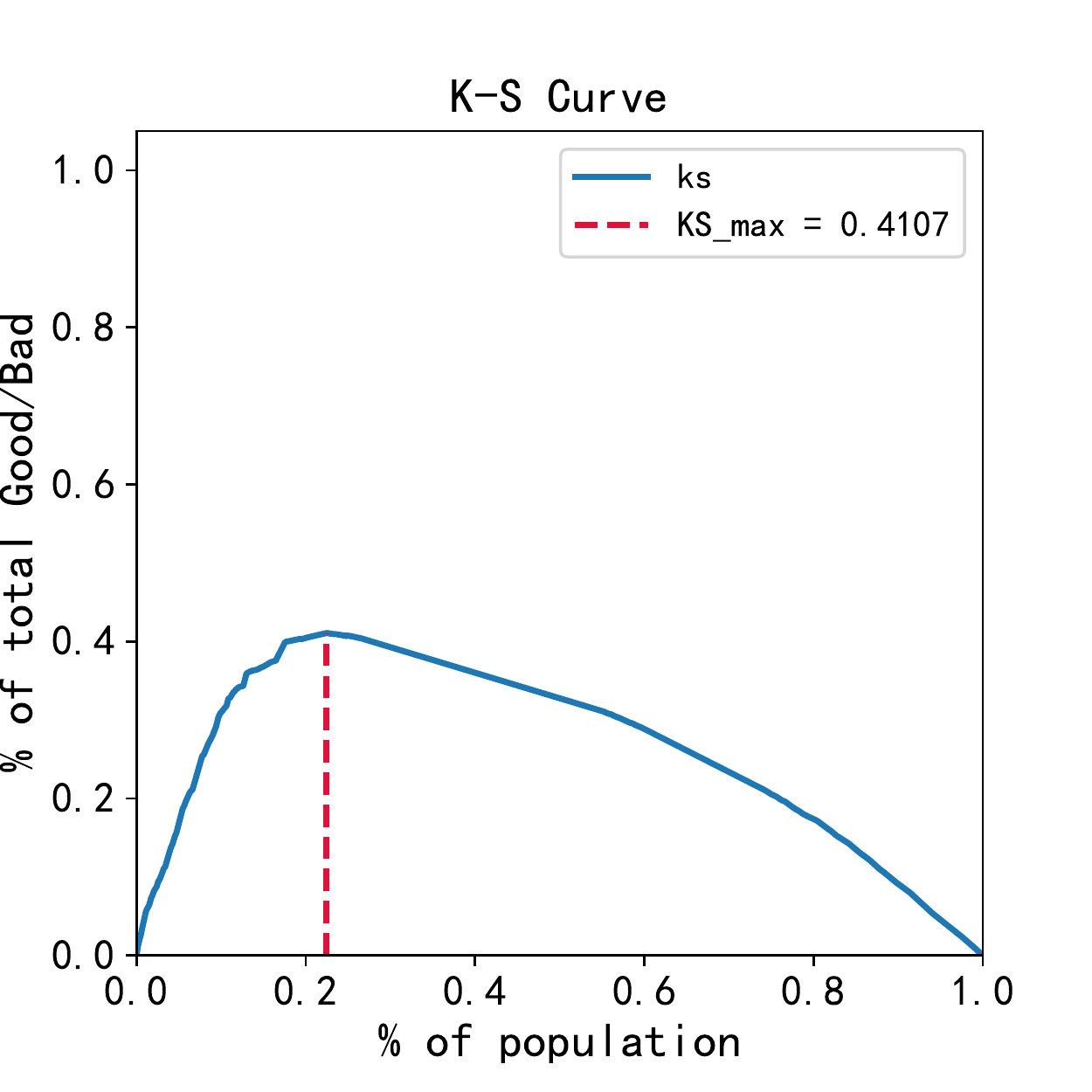}
\caption{ROC and KS curves for both training and testing parts using FL-LRBC.}
\label{fig:fl_measure} 
\end{figure}

The summary results of the scorecard model are shown in Table \ref{tab:plain2}. 
As expected, multiple input variables of positive sign are var0\_{woe}, var1\_{woe}, var3\_{woe}, var4\_{woe}, var6\_{woe}, var8\_{woe}, var9\_{woe}, var10\_{woe} and var13\_{woe} , respectively. Their coefficient values range from 0.19 to 1.99.

\begin{figure}[htbp]
\centering
\includegraphics[width=0.45\textwidth]{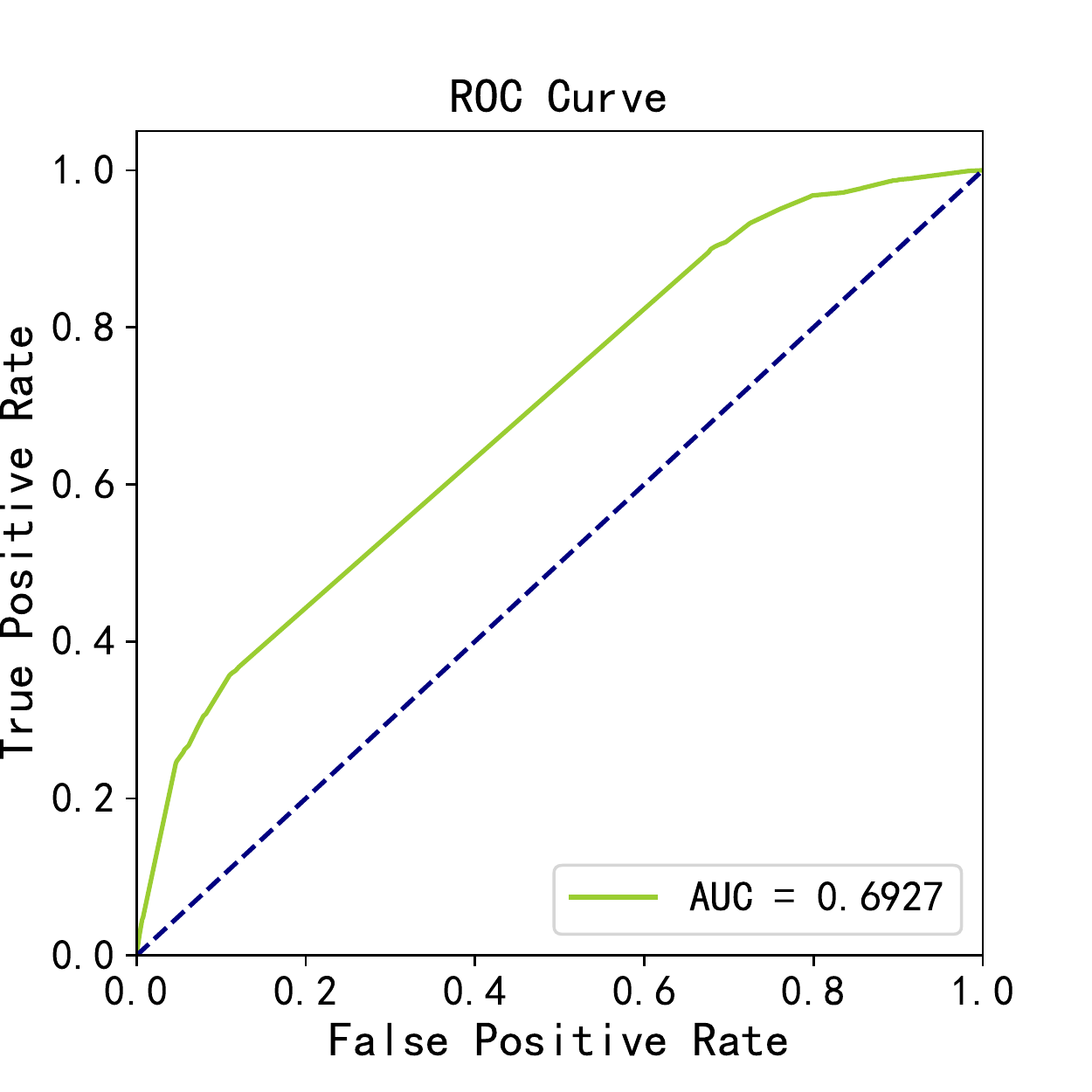}
\includegraphics[width=0.45\textwidth]{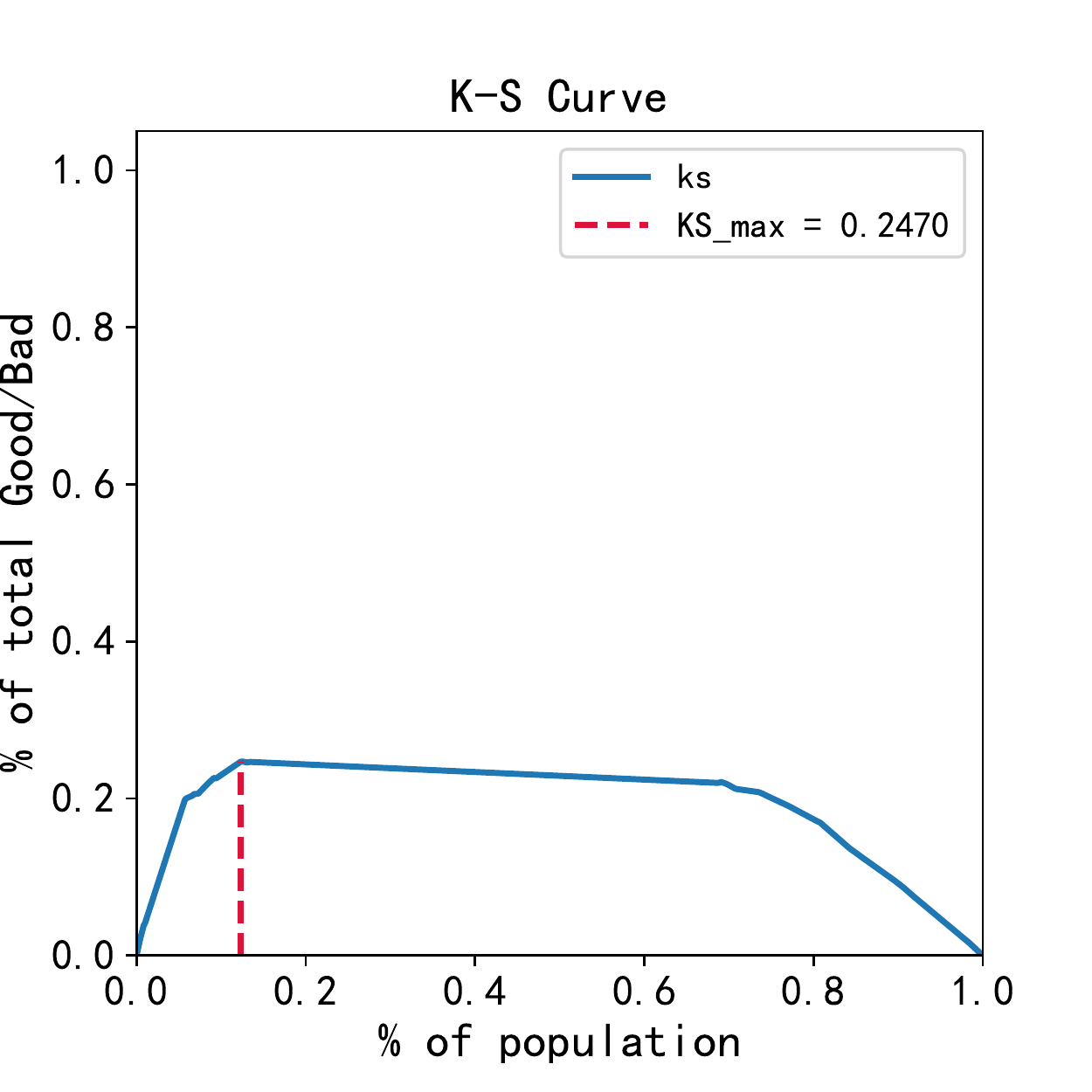}
\includegraphics[width=0.45\textwidth]{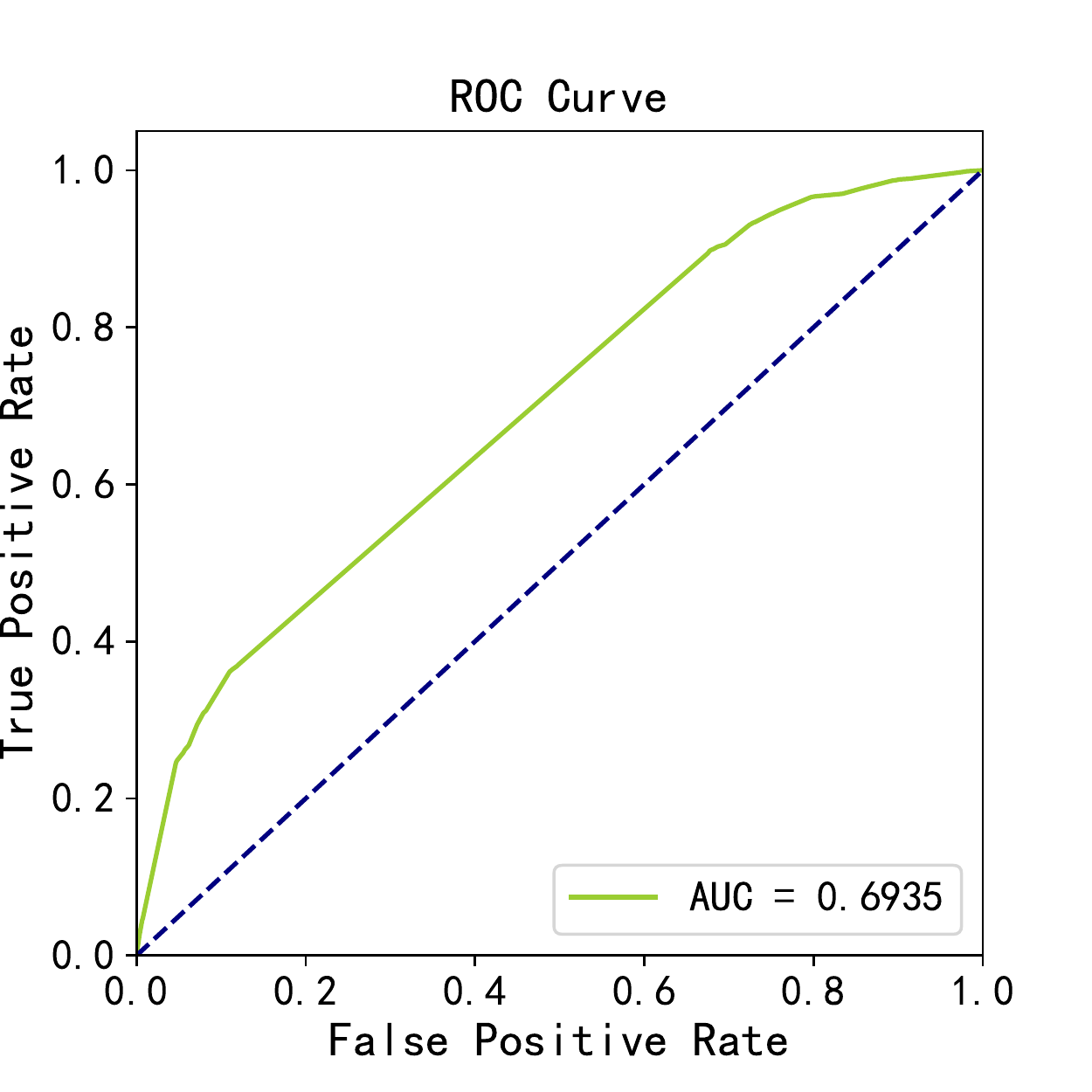}
\includegraphics[width=0.45\textwidth]{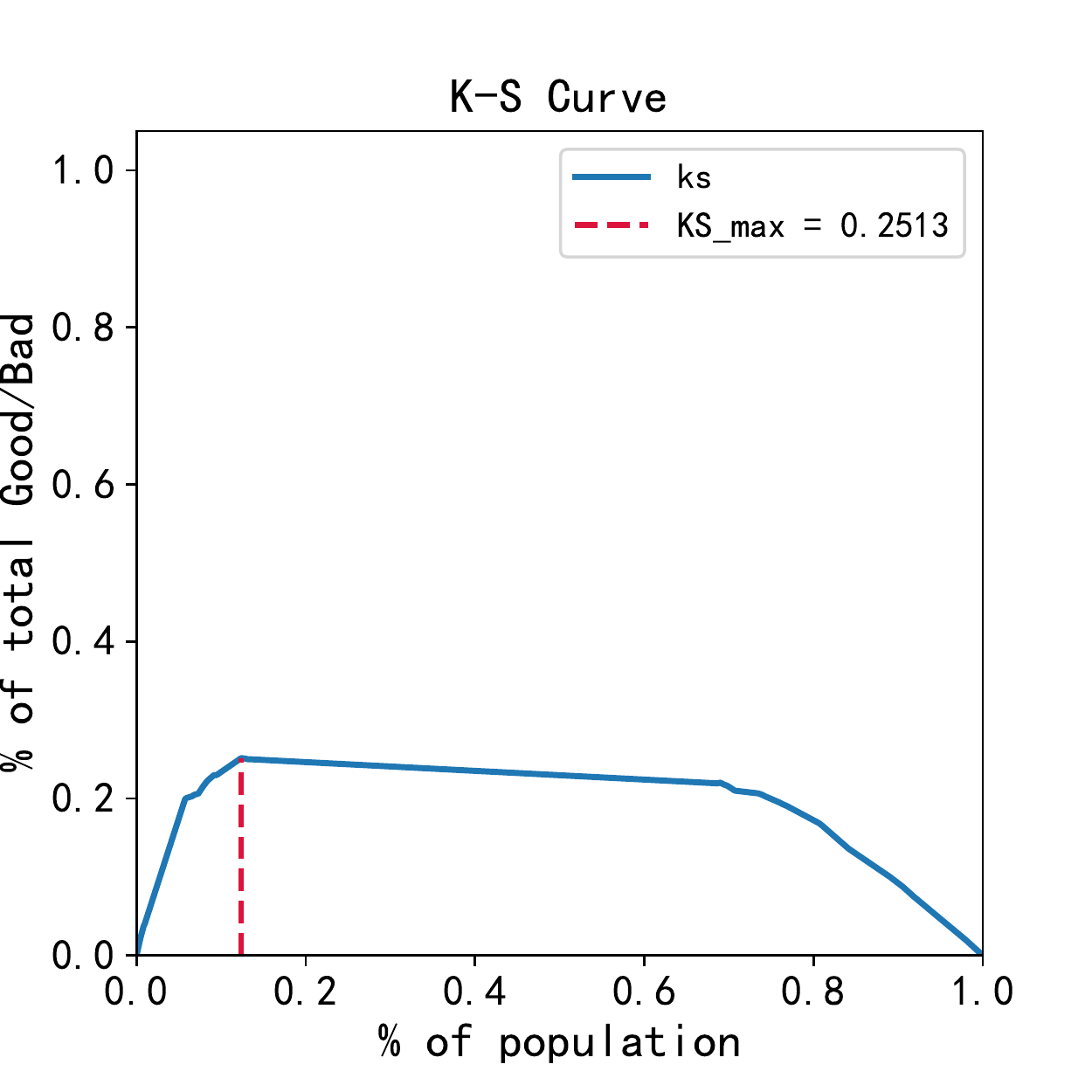}
\caption{ROC and KS curves for both training and testing parts using only BANK data.}
\label{fig:no_fl_measure} 
\end{figure}

Among these variables, six of them come from BANK
, which are characterized as identity traits, consumption grade, gross profits, behavioral preferences, channel activity. The rest three variables from 
CLOUD PAYMENT represent personal payment of apartment utilities (including electricity, water, gas and property charges) and mobile communication costs.

According to Figure \ref{fig:fl_measure} and Figure \ref{fig:no_fl_measure}, it is obvious that the performance in terms of AUC and KS statistics is remarkably improved due to data enrichment. Compared to the central bank’s scorecard using only BANK 
data, AUC of the FL-LRBC model has an increase of more than 10$\%$, reaching 0.76. The other metrics, KS statistic of 0.41 also has an evident improvement with raise by 60$\%$. 

Furthermore, cooperation between two agencies in GROUP
enriches the data set in terms of input features, 
leading that FL-LRBC based scorcard trends to be more stable and robust. 
As a result, with the usage of the FL-LRBC, 
default and non-default will be classified more effectively 
due to scorecard improvement.

Next, we will cooperate with more agencies such as security companies, insurance companies, trust companies, and E-commerce platforms in multiple business scenarios with the help of our FL-LRBC framework.

\section{Conclusion}\label{con}
In this manuscript, a projected gradient-based method in the FL framework for LR with bounded constraints (FL-LRBC) is proposed. 
Without the disclosure risk of data, an optimized interpretable scorecard, still based on the data from different data providers, can be obtained in a single training session. 
It not only avoids time-consuming feature-selecting and retraining process, but also ensures interpretability and robustness of the model. 
Compared with that without the data enrichment using the FL-LRBC, both AUC and KS statistics in evaluating model performance are highly improved.

\section*{Acknowledgements}
The authors gratefully acknowledge 
the colleagues from BANK and CLOUD PAYMENT in GROUP 
for our valuable discussions on business understanding and inspirations for the application design. The computing was executed on our Group's Data Haven (EDH), so we would like to express the deepest gratitude to the substantial help from EDH.

\bibliography{acml20}






\end{document}